\pdfoutput=1

\documentclass[11pt]{article}

\usepackage[]{acl}

\usepackage{times}
\usepackage{color}
\usepackage{latexsym}
\usepackage{graphicx}
\usepackage{amsmath}
\usepackage{booktabs}
\usepackage{multirow}
\usepackage{url}
\usepackage{amssymb}
\usepackage{enumitem}
\setitemize[1]{itemsep=0pt,partopsep=0pt,parsep=\parskip,topsep=0pt}
\usepackage{makecell}
\usepackage{float}
\usepackage{subfig}

\usepackage[T1]{fontenc}

\usepackage[utf8]{inputenc}

\usepackage{microtype}

\newcommand{\dogeticket}[1]{\includegraphics[width=#1\textwidth]{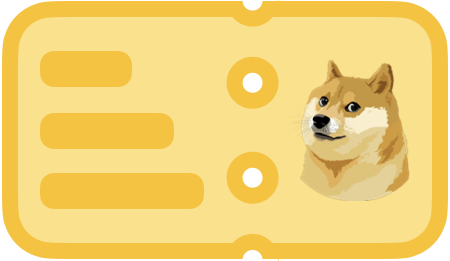}}
\title{\dogeticket{0.039}~Doge Tickets: Uncovering Domain-general Language Models by Playing Lottery Tickets}

\author{{Yi Yang}\textsuperscript{1}\Thanks{ Yi Yang and Chen Zhang contribute equally to this work, and the order is determined alphabetically.}, {Chen Zhang}\textsuperscript{1}\footnotemark[1], {Benyou Wang}\textsuperscript{2}, {Dawei Song}\textsuperscript{1}\Thanks{ Dawei Song is the corresponding author.} \\
   \textsuperscript{1}{Beijing Institute of Technology} \\
   \texttt{\{yang.yi,czhang,dwsong\}@bit.edu.cn} \\
   \textsuperscript{2}{The Chinese University of Hong Kong, Shenzhen} \\
   \texttt{wabyking@gmail.com}
}
\begin{document}
\maketitle
\begin{abstract}
Over-parameterized models, typically pretrained language models (LMs), have shown an appealing expressive power due to their small learning bias. However, the huge learning capacity of LMs can also lead to large learning variance. In a pilot study, we find that, when faced with multiple domains, a critical portion of parameters behave unexpectedly in a domain-specific manner while others behave in a domain-general one. Motivated by this phenomenon, we for the first time posit that domain-general parameters can underpin a domain-general LM that can be derived from the original LM. To uncover the domain-general LM, we propose to identify \textit{do}main-\textit{ge}neral parameters by playing lottery \textit{tickets} (dubbed \textit{doge tickets}). In order to intervene the lottery, we propose a \textit{domain-general score}, which depicts how domain-invariant a parameter is by associating it with the variance. Comprehensive experiments are conducted on the \textsc{Amazon}, \textsc{Mnli} and \textsc{OntoNotes} datasets. The results show that the \textit{doge tickets} obtains an improved out-of-domain generalization in comparison with a range of competitive baselines. Analysis results further hint the existence of domain-general parameters and the performance consistency of \textit{doge tickets}.\footnote{Code is available at \url{https://github.com/Ylily1015/DogeTickets}.}
\end{abstract}

\section{Introduction}

It is witnessed that more and more models become increasingly over-parameterized, typically pretrained language models (LMs). With the tremendous amounts of parameters, these models have shown an appealing expressive power ~\citep{DevlinCLT19,Liu19,RaffelSRLNMZLL20,Wang19}. While such LMs are enabled with small learning bias, they suffer from the limitation of large learning variance~\citep{NamkoongD17}, especially when faced with multiple domains.

\begin{figure}[t]
    \centering
    \includegraphics[width=0.43\textwidth]{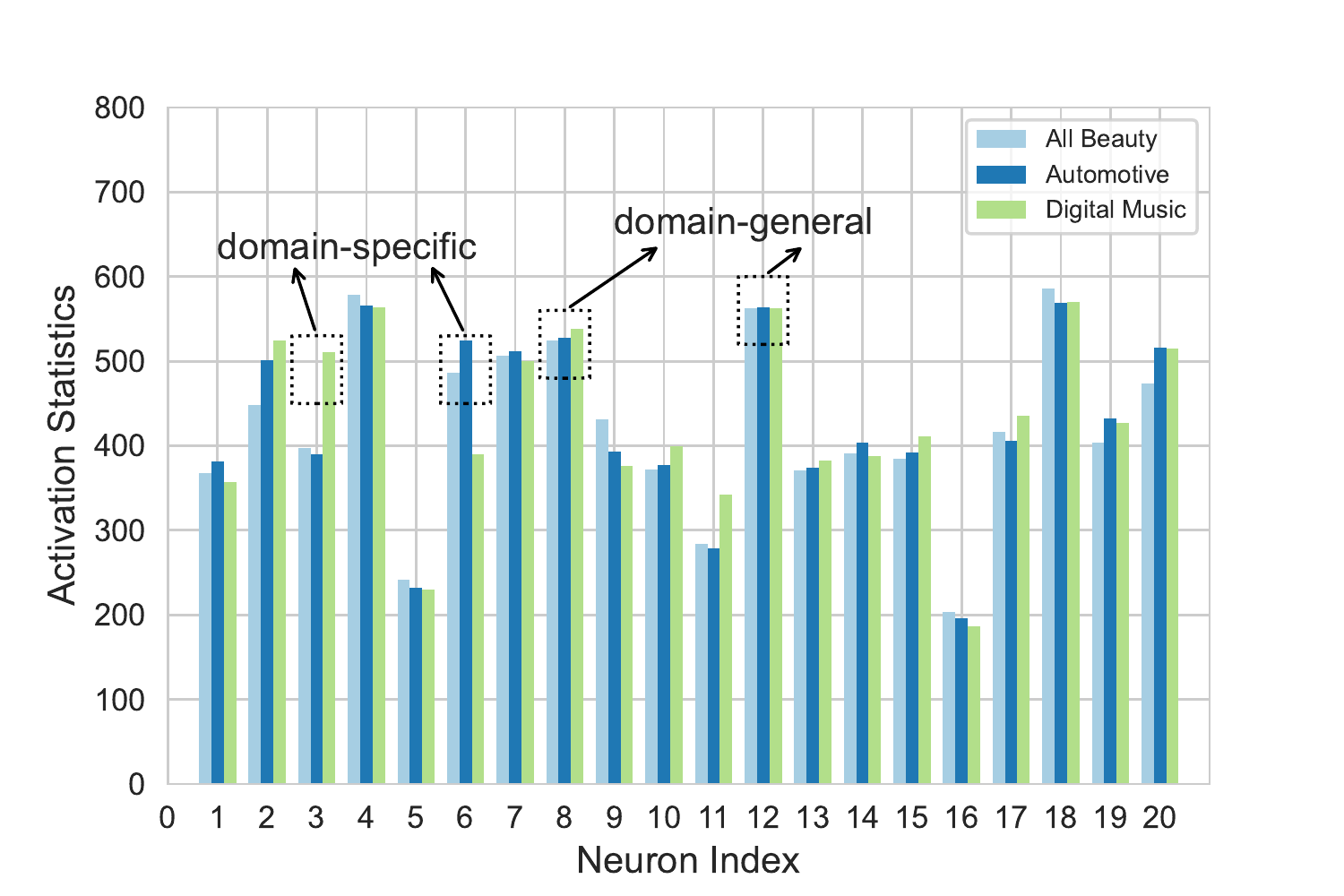}
    \caption{A pilot study on how sampled neurons (for \texttt{[CLS]} token) corresponding to the last feed-forward network (FFN) block of the finetuned \texttt{BERT}~\citep{DevlinCLT19} can be activated by training domains sampled from \textsc{Amazon}~\citep{NiLM19} via activation statistics~\citep{Bengio13}. Extended results are available at Appendix~\ref{appa}. Considering that GELU~\citep{HendrycksG16} activation is used, we view neurons with values larger than minus two as activated. A parameter is said to be specific to a domain or a group of domains if its associated neurons are particularly activated by input from these domains, while not so active in other domains. A parameter is said to be general to domains if its associated neurons are similarly activated by input across different domains. Both domain-specific and domain-general manners are observed.}
    \label{fig1}
\end{figure}

To show the consequence of such learning variance, we performed a pilot study on how different parameters of \texttt{BERT} behave over multiple domains (e.g., Digital Music, All Beauty, and Gift Cards). We find that a large portion of parameters are domain-specific, while others are domain-general. Figure~\ref{fig1} shows an illustrative example. Certain neurons corresponding to domain-specific parameters may be particularly activated by some domains but not so active in others. The domain-inconsistency can potentially lead to a deteriorated out-of-domain generalization capability, concerning how well a LM performs universally on any domains~\citep{NamkoongD17,Ye21}. In contrast, the neurons corresponding to domain-general parameters are similarly activated across different domains. 

Motivated by the phenomenon, we for the first time posit that, a domain-general LM that is underpinned by domain-general parameters as illustrated in Figure~\ref{fig1}, can be derived from the original LM, and that the domain-general LM would facilitate a better out-of-domain generalization. Specifically, inspired by lottery ticket hypothesis stating that a pruned model is capable of performing as expressive as the original over-parameterized model, we propose to identify \textit{do}main-\textit{ge}neral parameters by playing lottery \textit{tickets} (dubbed \textit{doge tickets})~\citep{FrankleC19} under the guidance of a domain-general score. The domain-general score describes how domain-invariant a parameter is, rather than how expressive as depicted by the commonly used expressive score~\citep{MichelLN19}. Driven by variance reduction~\citep{NamkoongD17}, the domain-general score associates with the invariance of a parameter by looking at not only the mean but also the variance of its expressive scores across different domains.

Comprehensive experiments are conducted on three widely used datasets: the \textsc{Amazon} sentiment classification dataset~\citep{NiLM19}, the \textsc{Mnli} language inference dataset~\citep{WilliamsNB18} and the \textsc{OntoNotes} named entity recognition dataset~\citep{Pradhan13}. The results demonstrate that the proposed \textit{doge tickets} owns a competitive out-of-domain generalization compared with an array of state-of-the-art baselines.

In-depth analyses are further carried out to double-check whether the domain-specific manner holds de facto and inspect whether the \textit{doge tickets} performs consistently. The results hint the existence of domain-general parameters and the performance consistency of \textit{doge tickets}.


\section{Background}

\subsection{Out-of-domain Generalization}

Pretrained LMs~\citep{DevlinCLT19,Liu19,RaffelSRLNMZLL20} have achieved a compelling performance on a range of downstream tasks where the training and test examples are identically distributed~\citep{Wang19}, thanks to the smaller learning bias brought by over-parameterization. Further exploration has also been carried out on whether LMs can have a good out-of-domain generalization~\citep{Hendrycks20}, where the training and test examples are distinguished from each other in terms of distributions (more specifically, they belong to different domains).

However, the performance gain is somewhat limited or even degraded, due to the fact that, on the one hand the over-parameterization brings small learning bias, but on the other hand it also brings large learning variance. Therefore, there is a large room for a better out-of-domain generalization of LMs~\citep{Ye21}. 

However, the issue with the large learning variance of pretrained LMs is yet to be systematically explored. In a pilot study (c.f. Section~\ref{sec3.1}), we find that under the setting of out-of-domain generalization, a number of parameters exhibit a domain-specific activation behavior, while others in a domain-general one. Motivated by this finding, we posit that domain-general parameters can underpin domain-general LMs. Therefore, provided that the training and test examples are separately sampled from training domains $\mathcal{D}$ and test domains $\mathcal{D}^{\prime}$, we aim to uncover a domain-general LM $\mathcal{M}^{\prime}$ from the original LM $\mathcal{M}$, and $\mathcal{M}^{\prime}$ is expected to have a better out-of-domain generalization ability.

It is also noteworthy that out-of-domain generalization is significantly different from domain adaptation, since domain adaptation has access to either labeled or unlabeled test examples.

\subsection{Lottery Ticket Hypothesis}

The lottery ticket hypothesis (LTH) states that a pruned model is capable of performing as good as the original over-parameterized model~\citep{FrankleC19}. LTH can be applied to LMs in various ways~\citep{Gale19,BrixBN20,ChenFC0ZWC20,MichelLN19,LiangZCJLHZC20} yet indicating a similar conclusion that \textit{lottery tickets can win}. Previous work discovers \textit{winning tickets} in LMs via either unstructured~\citep{FrankleC19,RendaFC20,ChenFC0ZWC20} or structured pruning techniques~\citep{MichelLN19,PrasannaRR20,ChenFC0ZWC20}. Unstructured pruning methods~\citep{Han15,Park17,Louizos18,Lee19} concentrate on pruning parameters at neuron level. In contrast, structured pruning methods~\citep{Li17,LuoL17,He17,He18} prune parameters at module level, leading to more feasibly accelerated LMs. In particular,~\citet{LiangZCJLHZC20} investigates the transition behavior of \textit{winning tickets} in LMs through shifting the sparsity levels, and has shown some slight performance gain with appropriate tickets (termed \textit{super tickets}) at a small sparsity level. 

Inspired by LTH, we propose to identify the domain-general parameters (in other words, remove domain-specific parameters) by playing lottery tickets (dubbed \textit{doge tickets}) under the guidance of a newly proposed domain-general score (c.f. Section~\ref{sec3.2}).

\subsection{Transformer Architecture}

A typical LM, say \texttt{BERT}~\citep{DevlinCLT19}, consists of a stack of transformer encoder layers~\citep{VaswaniSPUJGKP17}. Each of these layers contains two blocks: a multi-head self-attention block (MHA) and a feed-forward network block (FFN), with a residual connection and a normalization layer around each. In the following description, we omit the details of residual connections, normalization layers and bias terms for brevity.

Assuming there are $n$ independent heads in each MHA block and the $i$-th head is parameterized by $\mathbf{W}_{Q}^{(i)}$, $\mathbf{W}_{K}^{(i)}$, and $\mathbf{W}_{V}^{(i)}\in\mathbb{R}^{d\times d_h}$, then an MHA block can be depicted as:
\begin{equation}\nonumber
\mathbf{H}^{(i)}(\mathbf{X})=\text{Att}(\mathbf{X}\mathbf{W}_{Q}^{(i)},\mathbf{X}\mathbf{W}_{K}^{(i)},\mathbf{X}\mathbf{W}_{V}^{(i)}) 
\end{equation}
where $\mathbf{X}\in\mathbb{R}^{l\times d}$ represents $d$-dimensional vectors that stand for $l$ sequential input token representations. The intermediate outputs of all parallel heads are further processed and summed to produce the final output from the MHA block:
\begin{equation}\nonumber
\mathbf{Z}=\text{MHA}(\mathbf{X})=\sum_{i=1}^{n}\mathbf{H}^{(i)}(\mathbf{X})\mathbf{W}_O^{(i)}
\end{equation}
where $\mathbf{W}_O^{(i)}\in\mathbb{R}^{d_h\times d}$ is the output linear layer for the $i$-th head.

Likewise, we denote the output of the FFN block, which is composed of two linear layers with proper activation in-between as:
\begin{equation}\nonumber
\mathbf{X}^{\prime}=\text{FFN}(\mathbf{Z})=\mathbf{W}_2 \text{GELU}(\mathbf{W}_1\mathbf{Z})
\end{equation}
where $\mathbf{W}_1\in\mathbb{R}^{d\times d_i}$ and $\mathbf{W}_2\in\mathbb{R}^{d_i\times d}$ correspond to the two layers.

Our work exactly targets at improving out-of-domain generalization through uncovering a domain-general \texttt{BERT} from the original \texttt{BERT} by playing lottery tickets.

\section{Identifying Doge Tickets}

\begin{figure*}[t]
    \centering
    \includegraphics[width=0.97\textwidth]{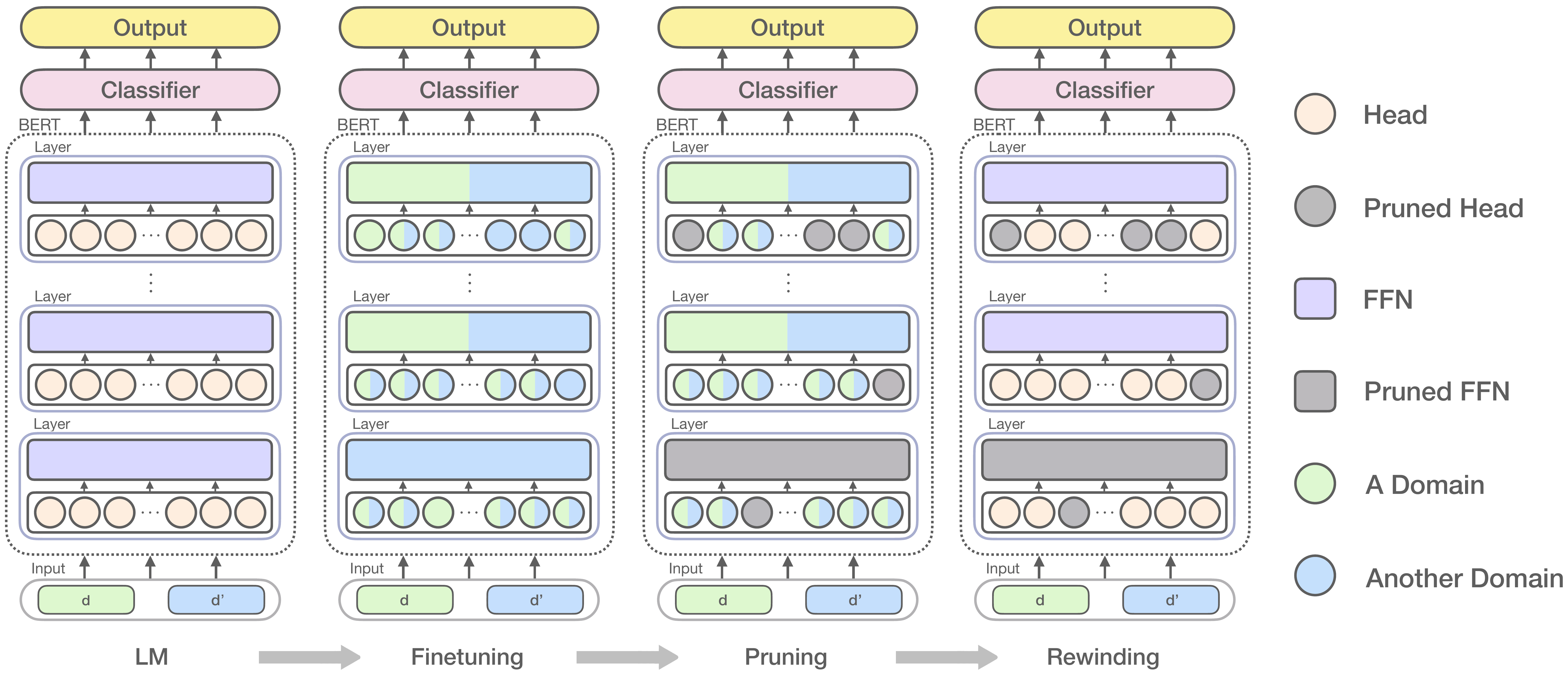}
    \caption{\textit{first finetuning, then pruning, finally rewinding} paradigm when playing lottery tickets. It is found that parameters can behave in either a domain-specific or domain-general manner after finetuning. Pruning is conducted to prune the domain-specific parameters of the LM. Rewinding necessarily sets the pruned LM to the initial point of finetuning and trains it again.}
    \label{fig2}
\end{figure*}

\subsection{Uncovering Domain-general LM}
\label{sec3.1}

LMs are perceived to bring large learning variance due to over-parameterization. Accordingly, we suspect that the variance to some extent showcases itself within the LMs. Under the setting of out-of-domain generalization, we conduct a pilot study to examine how neurons activation~\citep{Bengio13} can vary from a domain to another, as in Figure~\ref{fig1}. 

In the study, the activation statistics towards three training domains of a sampled neuron (for \texttt{[CLS]} token) from each FFN block are attained by 1) feeding the examples sampled from these domains into the fined-tuned \texttt{BERT}; and 2) counting the accumulated times of activation. Recapping the results in Figure~\ref{fig1}, we can find that quite a few neurons, or parameters without loss of generality, unexpectedly behave in a domain-specific manner, which may well cause the unsatisfying out-of-domain generalization performance. Yet, other parameters may behave in a domain-general manner.

Motivated by the phenomenon, we hypothesize that domain-general parameters may essentially underpin a domain-general \texttt{BERT} that can be derived from the original \texttt{BERT}. This makes it possible to uncover a domain-general \texttt{BERT} by simply identifying domain-general parameters. 

Witnessing the success of LTH that states a well-performing pruned model can be uncovered by playing lottery tickets~\citep{FrankleC19}, we seek an effective solution customized for the domain-general \texttt{BERT} from intervening lottery tickets (dubbed \textit{doge tickets}).

\subsection{Playing Lottery Tickets}
\label{sec3.2}

The identification of tickets typically follows a \textit{first finetuning, then pruning, finally rewinding} paradigm~\citep{FrankleC19}. Following recent advances~\citep{MichelLN19,PrasannaRR20,LiangZCJLHZC20}, we would like to apply structured pruning to a LM to identify \textit{doge tickets} by pruning MHA heads and FFN blocks are concerned. The most basic reason sits in that unstructured pruning requires maintaining a global pool of scores for all parameters, leading to a predominant memory overhead. Although we have only discovered domain-specific manner at neuron level in our pilot study (i.e. Section~\ref{sec3.1}), we believe the same can apply at module level. For this purpose, the procedure is illustrated in Figure~\ref{fig2}, including the following steps: 1) finetuning: train the LM on a specific task; 2) pruning: prune parameters of the LM to a target sparsity under the guidance of the domain-general scores computed with the  trained LM; 3) rewinding: set the pruned LM to the initial point of finetuning and train it on the task again.

Previous work identifies \textit{winning tickets} by referring to the \textbf{expressive scores} of parameters~\citep{MichelLN19}. While \textit{winning tickets} can discover more light-weighted and possibly more expressive LMs, we argue that they are not suitable for identifying \textit{doge tickets} since the variance has not been taken into consideration. Thus, we intervene the playing of lottery tickets by referring to the \textbf{domain-general scores} which take into consideration the mean and variance of expressive scores across domains. 
\paragraph{Expressive Scores.}

The expressive scores are approximated by first masking the parameterized elements of the finetuned LM.  $\xi^{(i)}$ and $\nu$  generally denote the mask variables respectively for an MHA head and an FFN block~\citep{PrasannaRR20}, such that:
\begin{equation}\nonumber
\begin{aligned}
^{\circ}\text{MHA}(\mathbf{X})&=\sum_{i=1}^{n}\xi^{(i)}\mathbf{H}^{(i)}(\mathbf{X})\mathbf{W}_O^{(i)}\\
^{\circ}\text{FFN}(\mathbf{Z})&=\nu\mathbf{W}_2 \text{GELU}(\mathbf{W}_1\mathbf{Z})
\end{aligned}
\end{equation}
where we initialize the mask variables $\xi^{(i)}=1$ and $\nu=1$ for the corresponding MHA head and FFN block to preserve the original LM.

Then the expected absolute gradient over all training data for the MHA head and FFN block gives the expressive scores:
\begin{equation}\nonumber
\begin{aligned}
\mathbb{I}_{\text{MHA}}^{(i)}&=\mathbb{E}_{(x,y)\sim\mathcal{D}}\left|\frac{\partial\mathcal{L}(x,y)}{\partial\xi^{(i)}}\right|\\
\mathbb{I}_{\text{FFN}}&=\mathbb{E}_{(x,y)\sim\mathcal{D}}\left|\frac{\partial\mathcal{L}(x,y)}{\partial \nu}\right|\\
\end{aligned}
\end{equation}
where $(x,y)$ is a data point and $\mathcal{L}$ is the loss function. $\mathbb{E}$ represents expectation.

The absolute value of gradient for a mask indicates how large the impact of masking the corresponding element is, thus implying how expressive an element is. Intuitively, a less expressive element should be pruned preferentially.

\paragraph{Domain-general Scores.}

After obtaining expressive scores in all domains, the domain-general scores take the mean and further the variance of expressive scores across domains. The domain-general scores can be formulated as:
\begin{equation}\nonumber
\begin{aligned}
\mathbb{I}_{\text{MHA}}^{(i)\prime}&=\mathbb{E}_{d\sim\mathcal{D}}\mathbb{E}_{(x,y)\sim d}\left|\frac{\partial\mathcal{L}(x,y)}{\partial\xi^{(i)}}\right|\\
&-\lambda\mathbb{V}_{d\sim\mathcal{D}}\mathbb{E}_{(x,y)\sim d}\left|\frac{\partial\mathcal{L}(x,y)}{\partial\xi^{(i)} }\right|\\
\mathbb{I}_{\text{FFN}}^{\prime}&=\mathbb{E}_{d\sim\mathcal{D}}\mathbb{E}_{(x,y)\sim d}\left|\frac{\partial\mathcal{L}(x,y)}{\partial \nu}\right|\\
&-\lambda\mathbb{V}_{d\sim\mathcal{D}}\mathbb{E}_{(x,y)\sim d}\left|\frac{\partial\mathcal{L}(x,y)}{\partial\nu}\right|
\end{aligned}
\end{equation}
where $(x,y)$ is a data point from the domain $d$.  The domain-general score measures the balance between the mean and variance of expressive scores across domains. $\lambda$ is adopted to quantify the trade-off.As suggested by~\citet{MolchanovTKAK17}, we also normalize the expressive scores of MHA heads in each layer with $\ell_{2}$ norm. 

We gradually increase sparsity level by pruning the elements with lowest domain-general scores, therefore producing LMs with different sparsity levels. Rewinding is applied to these LMs by setting the remaining parameters' values to what they were in the early stage of finetuning~\citep{FrankleD0C20} (i.e., the initialization point for finetuning). We expect the LM with remaining parameters can achieve better out-of-domain generalization, thus regarding them as \textit{doge tickets}.


\begin{table*}[ht]
\centering
\resizebox{0.97\textwidth}{!}{

\begin{tabular}{cccccc}
    \toprule
    \textbf{Dataset} & $\mathcal{D}$ & \textbf{\#train.} & \textbf{\#dev.} & $\mathcal{D}^{\prime}$ & \textbf{\#test} \\ 
    \midrule
    \textsc{AmazonA} & \makecell[c]{\{All Beauty, Automotive, Digital Music,\\ Gift Cards\}} & \multirow{6}{*}{5,400}      & \multirow{6}{*}{600}     & \{Industrial and Scientific, Movies, Software\} & \multirow{6}{*}{6,000} \\
    \cmidrule{1-2}\cmidrule{5-5}
    \textsc{AmazonB} & \makecell[c]{\{All Beauty, Industrial and Scientific, \\Movies, Software\}} &       &      & \{Automotive, Digital Music,	Gift Cards\} &  \\
    \cmidrule{1-2}\cmidrule{5-5}
    \textsc{AmazonC} & \makecell[c]{\{Digital Music, Gift Cards, Movies,\\ Software\}} &       &      & \{All Beauty, Automotive, Industrial and Scientific\} &  \\
    \midrule
    \textsc{Mnli}    & \makecell[c]{\{Fiction, Government, Slate,\\ Telephone, Travel\}} & 78,540      & 1,963     & \makecell[c]{\{Face to Face, Letters, Nine,\\ Oup, Verbatim\}}       & 1,966     \\
    \midrule
    \textsc{OntoNotes}    & \makecell[c]{\{Broadcast Conversation, Broadcast
News,
\\Magazine, Newswire\}} & 16,111      & 2,488     & \makecell[c]{\{Telephone Conversation, Web Data\}}       & 1,837     \\
    \bottomrule
\end{tabular}}
\caption{Statistics of datasets. \textbf{\#train.}, \textbf{\#dev.}, and \textbf{\#test} indicate average number of training, development, and test examples per domain.}
\label{tab1}
\end{table*}

\section{Experiments}

\subsection{Datasets}

We conduct our experiments on the \textsc{Amazon} sentiment classification dataset~\citep{NiLM19}, the \textsc{Mnli} language inference dataset~\citep{WilliamsNB18} and the \textsc{OntoNotes} named entity recognition dataset~\citep{Pradhan13}. They are described as follows:
\begin{itemize}[leftmargin=*]
\item \textsc{Amazon} originally contains reviews from a variety of product categories. We randomly select 7 product categories as domains and sample 6,000 reviews from each domain. Since each review is associated with a score ranging from 1 to 5 stars, we derive a 3-way sentiment labels from the score by viewing a score larger than 3 as positive, a score smaller than 3 as negative, and a score of 3 as neutral. Then we randomly select 4 out of 7 domains as training set ($\mathcal{D}$) and the rest 3 as test set ($\mathcal{D}^{\prime}$). For development purpose, we further sample 1/10 from each training domain as gathered development set. In this way, we can construct an out-of-domain dataset. This process is repeated for three times, resulting in a total of 3 out-of-domain datasets, denoted as \textsc{AmazonA}, \textsc{AmazonB}, and \textsc{AmazonC} respectively. 
\item \textsc{Mnli} covers a range of genres of spoken and written sentence pairs with textual entailment information. We strictly follow its original data split and use the mismatched (i.e., different genres from those in the training set) development set of the dataset as out-of-domain test set, giving us 5 training domains and 5 test domains.
\item \textsc{OntoNotes} consists of annotated named entities types from 6 genres. We randomly choose 4 domains as training set and the rest 2 as test set.
\end{itemize}

The detailed information and statistics of the constructed datasets are presented in Table~\ref{tab1}.

\subsection{Baselines \& Implementation}

We directly finetune the pretrained \texttt{BERT}~\citep{DevlinCLT19} (in fact, \texttt{BERT-base}) without pruning to obtain a solid baseline for out-of-domain generalization measure, and further carry out comparisons among the following models:

\begin{itemize}[leftmargin=*]
\item \texttt{BERT} w/ invariant risk minimization (\texttt{IRM}) applies a domain-invariant classifier~\citep{Arjovsky19} to the \texttt{BERT}, serving as a strong baseline for out-of-domain generalization.
\item \texttt{BERT} w/ \textit{random tickets} rewinds on a set of \textit{random tickets} randomly sampled from the structured modules (MHA heads and FFN blocks).
\item \texttt{BERT} w/ \textit{winning tickets} rewinds on a set of \textit{winning tickets} chosen according to the expressive scores over training data. We identify the tickets through pruning the \texttt{BERT} to an expected sparsity.
\item \texttt{BERT} w/ \textit{doge tickets} rewinds on a set of \textit{doge tickets} chosen according to the domain-general scores over training data, which is akin to the procedure of \texttt{BERT} w/ \textit{winning tickets}.
\end{itemize}

We use AdamW~\citep{LoshchilovH19} as the optimizer in our experiments. The learning rate is searched within \{1, 2, 3, 5\}$\times$10\textsuperscript{-5} and the batch size is searched within \{16, 32\}. Both training and rewinding last no longer than 10 epochs, with an early-stopping.

\subsection{Evaluation Metrics}

For out-of-domain generalization, models are learned from training domains and evaluated on ever unseen test domains. For both finetuning and rewinding, we select the checkpoints based on the performance on the development set, and test these checkpoints on the test set. The averaged Accuracy and the average F1-score over test domains are adopted as evaluation metrics during our experiments.

\subsection{Main Comparison}

\begin{table*}[ht]
\centering
\resizebox{0.97\textwidth}{!}{
\begin{tabular}{lccccccc}
    \toprule
    \multirow{3}{*}{\textbf{Model}} & \multicolumn{5}{c}{\textbf{Datasets}} & \multirow{3}{*}{\makecell[c]{\textbf{Average} \\\textbf{Score}}} &  \multirow{3}{*}{\makecell[c]{\textbf{Average} \\\textbf{Sparsity}}} \\
    \cmidrule{2-6}
    & \textsc{AmazonA} & \textsc{AmazonB} & \textsc{AmazonC} & \textsc{Mnli} & \textsc{OntoNotes} & \\ 
    \cmidrule{2-6}
    & \textbf{Acc}    & \textbf{Acc}  & \textbf{Acc}    & \textbf{Acc} & \textbf{F1} & \\
    \midrule
    \texttt{BERT}      & 69.8 & 72.6 & 69.6 & 84.8 & 57.2 & 70.8  & 0.0\% \\
    \quad w/ \texttt{IRM} & 70.4 & 72.5 & 70.7 & 84.3 & 56.3 & 70.8 & 0.0\% \\
    \quad w/ \textit{random tickets} & 71.4 & 73.3 & 70.1 & 84.6 & 57.9 & 71.5 & 12.8\%  \\
    \quad w/ \textit{winning tickets} & 70.9 & 73.7 & 71.3 & 84.8 & 57.9 & 71.7 & 17.5\%  \\
    \quad w/ \textit{doge tickets} & \textbf{71.7} & \textbf{73.8} & \textbf{72.2} & \textbf{85.0} & \textbf{58.5} & \textbf{72.2} & 15.0\% \\
    \bottomrule
\end{tabular}}
\caption{Main comparison results in percentage. The best results on datasets are \textbf{boldfaced}. Average Score is the average metric over used datasets. Average Sparsity is the average sparsity to achieve best out-of-domain generalization among all sparsity levels over used datasets.}
\label{tab2}  
\end{table*}

Table~\ref{tab2} shows the main comparison results on all constructed datasets. As we can see, \texttt{BERT} w/ \textit{doge tickets} certainly generalizes better than \texttt{BERT}. Specifically, the accuracy improvements over \texttt{BERT} brought by \textit{doge tickets} are 1.9\% on \textsc{AmazonA}, 1.2\% on \textsc{AmazonB}, 2.6\% on \textsc{AmazonC}, 0.2\% on \textsc{Mnli} and 1.3\% on \textsc{OntoNotes}. Further, compared to the competitive out-of-domain generalization baseline, \texttt{BERT} w/ \texttt{IRM}, our model \texttt{BERT} w/ \textit{doge tickets} also achieves a performance improvement, which illustrates that domain-specific parameters would better be pruned instead of regularized. Finally, \texttt{BERT} w/ \textit{doge tickets} gains 0.5\% on average over that with \textit{winning tickets}, implying that domain variance should be considered as a significant part when playing the lotteries. Surprisingly, \texttt{BERT} w/ \textit{random tickets} can be competitive with \texttt{BERT} w/ \textit{winning tickets}, suggesting the sub-optimality of using \textit{winning tickets} for out-of-domain generalization. Meanwhile, we notice that \textit{doge tickets} favors a comparably smaller sparsity. This observation can be seen as an evidence that, with the variance, we should cautiously prune parameters.

We also provide results for \texttt{BERT-large} in Table~\ref{tab3}, which show that \textit{doge tickets} can be applied to larger LMs and achieve a performance gain. 

\begin{table}[ht]
    \centering
    \resizebox{0.37\textwidth}{!}{
    \begin{tabular}{lcc}
    \toprule
    \multirow{3}{*}{\textbf{Model}} & \textbf{Datasets} & \multirow{3}{*}{\makecell[c]{\textbf{Average} \\\textbf{Sparsity}}} \\
    \cmidrule{2-2}
    & \textsc{AmazonA} \\
    \cmidrule{2-2}
    & \textbf{Acc} \\
    \midrule
    \texttt{BERT-large} & 73.1 & 0.0\% \\
    \quad w/ \texttt{IRM} & 73.5 & 0.0\% \\
    \quad w/ \textit{winning tickets} & 74.0 & 15.0\% \\
    \quad w/ \textit{doge tickets} & \textbf{74.3} & 15.0\% \\
    \bottomrule
    \end{tabular}}
    \caption{Extended comparison results in percent. Larger LMs are used.}
    \label{tab3}
\end{table}

\section{Analyses}

\subsection{Sensitivity to Learning Variance}

\begin{figure}[ht]
    \centering
    \includegraphics[width=0.43\textwidth]{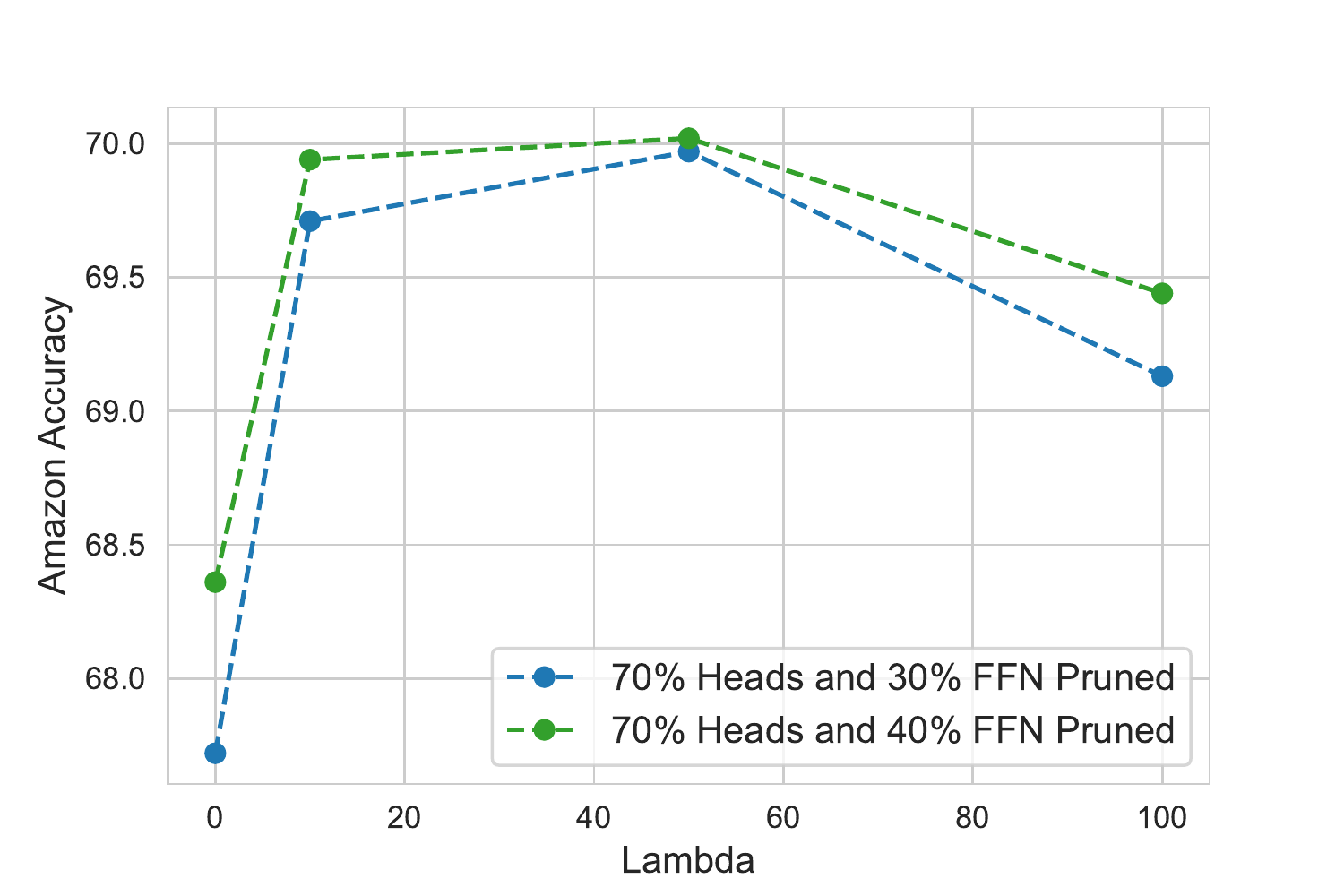}
    \caption{\textit{doge tickets} on \textsc{AmazonA} under various $\lambda$ values with two sparsity levels.}
    \label{fig5}
\end{figure}

To investigate the effect of learning variance for out-of-domain generalization, Figure~\ref{fig5} shows the results from different settings, \{0, 10, 50, 100\}, of the variance coefficient $\lambda$  on \textsc{AmazonA}. 

The results suggest that it is necessary to take into account the variance when quantifying domain-general scores. The model performance first increases till a certain $\lambda$ and then decreases, hinting that we can count the variance as much as possible on the same sparsity level to increase model generalization. 

\subsection{Impact of Training Domains}

One may well note that the performance improvement of \textit{doge tickets} is not significant on \textsc{Mnli}. We conjecture this is subject to the number of training domains. To explore whether training domain number matters, we conduct experiments by randomly selecting domains from the original 5 in \textsc{Mnli} (\textsc{Mnli}-5) as training domains, denoted as \textsc{Mnli}-4 and \textsc{Mnli}-3.

\begin{figure*}[ht]
    \centering
    \includegraphics[width=0.93\textwidth]{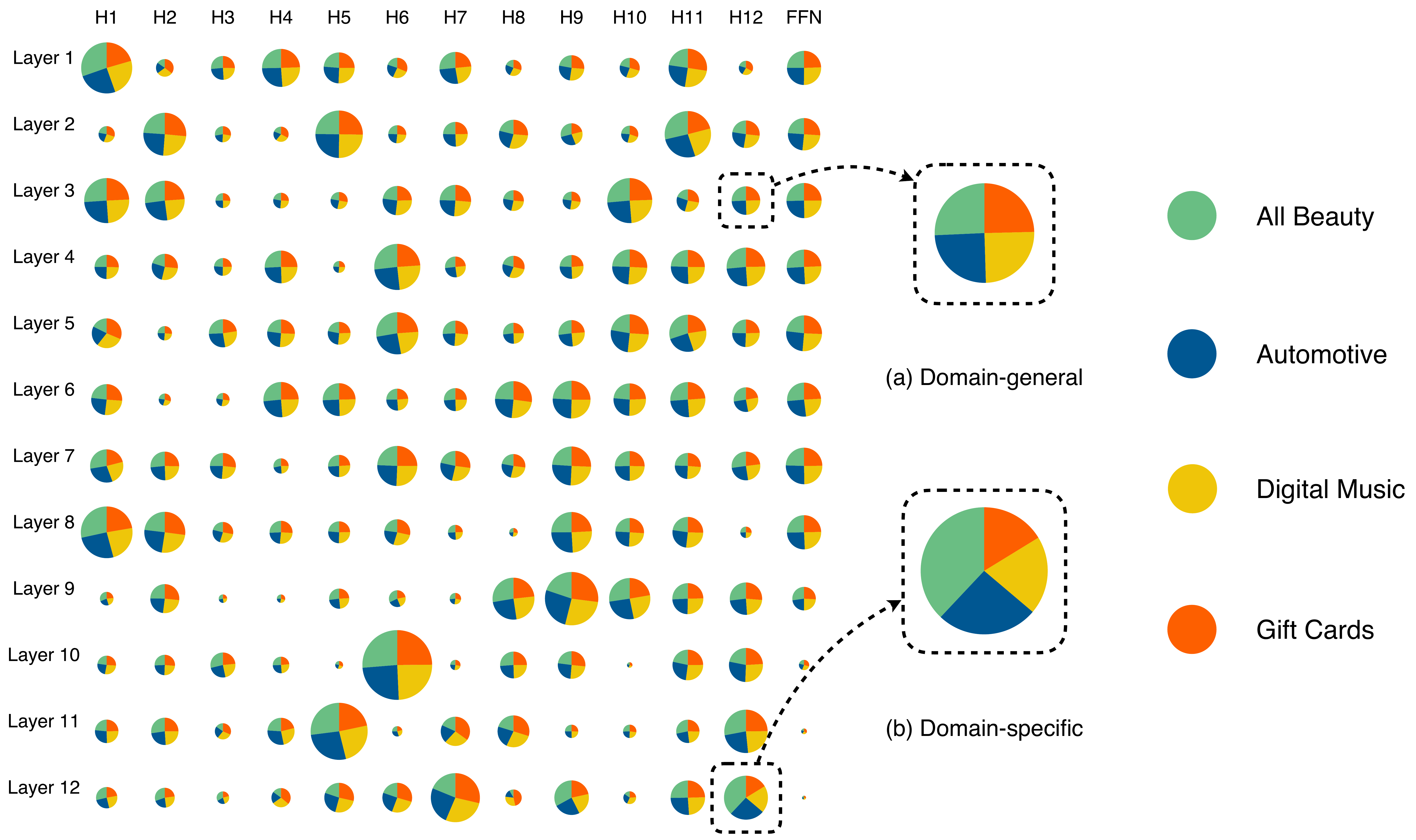}
    \caption{Illustration of expressive scores across domains. Each pie represents a parameterized element (either an MHA head or FFN block). The mean is measured by the radius of a pie. We use 4 distinguished colors to represent domains, whose details are shown in legend. The variance is measured by the proportion of each color in a pie.}
    \label{fig3}
\end{figure*}

As shown in Table~\ref{tab4}, we observe that \texttt{BERT} w/ \textit{doge tickets} discovers a larger generalization margin, when fewer (i.e., 3 and 4) training domains are used. That is, when using a smaller number of training domains, the impact of domain-specific (or domain-general) parameters on generalization results becomes more significant. According to the existing literature and intuition~\citep{Hendrycks20}, more training domains can help LMs to extrapolate to unseen domains to some extent, thereby achieving a better out-of-domain generalization than the use of less training domains. 

\begin{table}[ht]
\centering
\resizebox{0.45\textwidth}{!}{
\begin{tabular}{lcccc}
    \toprule
    \multirow{3}{*}{\textbf{Model}} & \multicolumn{3}{c}{\textbf{Datasets}} & \multirow{3}{*}{\makecell[c]{\textbf{Average} \\\textbf{Sparsity}}} \\
    \cmidrule{2-4} 
    & \textsc{Mnli}-5 & \textsc{Mnli}-4     & \textsc{Mnli}-3 & \\
    \cmidrule{2-4}
    & \textbf{Acc}    & \textbf{Acc}    & \textbf{Acc}  & \\
    \midrule
    \texttt{BERT}     & 84.8 &84.2 & 83.0 & 0.0\% \\
    \quad w/ \textit{winning tickets} & 84.8 &84.3 & 83.3 & 8.7\%  \\
    \quad w/ \textit{doge tickets} & 85.0 &84.5 & 83.6 & 5.3\% \\
    \qquad $\Delta$ & 0.2 &0.3 & 0.6 & -- \\
    \bottomrule
\end{tabular}}
\caption{Results in percentage on \textsc{Mnli} with different training domain numbers. $\Delta$ means generalization margin.}
\label{tab4}  
\end{table}

\subsection{Existence of Domain-specific Manner}
\label{existence}

While we have found that a critical portion of parameters behave in a domain-specific manner through our pilot observation over intermediate activation distributions within the LM, we would like to verify the finding via intuitively visualizing expressive scores of parameters across domains. We use \textsc{AmazonA} and show the expressive scores of parameterized elements across different domains in Figure~\ref{fig3}.

Based on the mean and variance of expressive scores across domains, the parameters can be divided into 4 types: (1) high mean with high variance (HMHV); (2) high mean with low variance (HMLV); (3) low mean with high variance (LMHV); and (4) low mean with low variance (LMLV). It is obvious that there exists quite a number of elements with small radius, showing they play a less important role in the LM. This kind of parameters are exactly LMHV and LMLV parameters. If we merely consider this, \textit{winning tickets} seems already a promising choice. However, more importantly, some elements are indeed behave unexpectedly in a domain-specific manner if we look at the proportions of different colors within pies, i.e., LMHV and HMHV parameters, suggesting the need of considering variance as in \textit{doge tickets}.

To highlight, LMLV and HMLV elements are actually applicable across different domains, such as Figure~\ref{fig3} (a). On the other hand, LMHV and HMHV ones are dedicated to a domain or two, for example ``All Beauty'' domain as Figure~\ref{fig3} (b) shows. So \textit{doge tickets} chooses to prune LMHV parameters in the first place yet HMLV in the last, leading to domain-general LMs.

\subsection{Consistency with Varying Sparsity Levels}

\begin{figure}[]
	\centering
	\subfloat[\texttt{\small BERT-base}]{\includegraphics[width=0.3\textwidth]{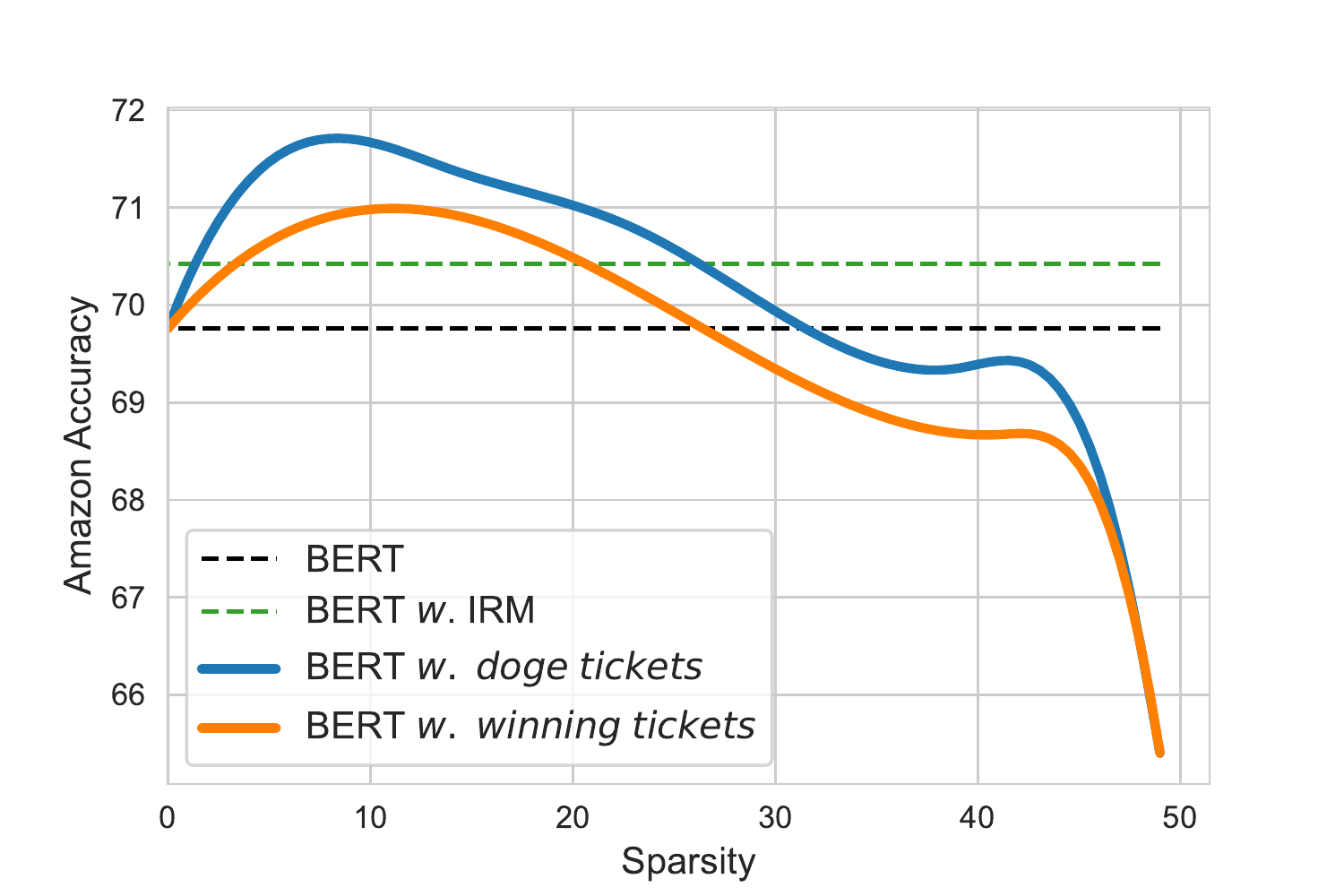}
	\label{fig4a}}\\
	\subfloat[\texttt{\small BERT-large}]{\includegraphics[width=0.3\textwidth]{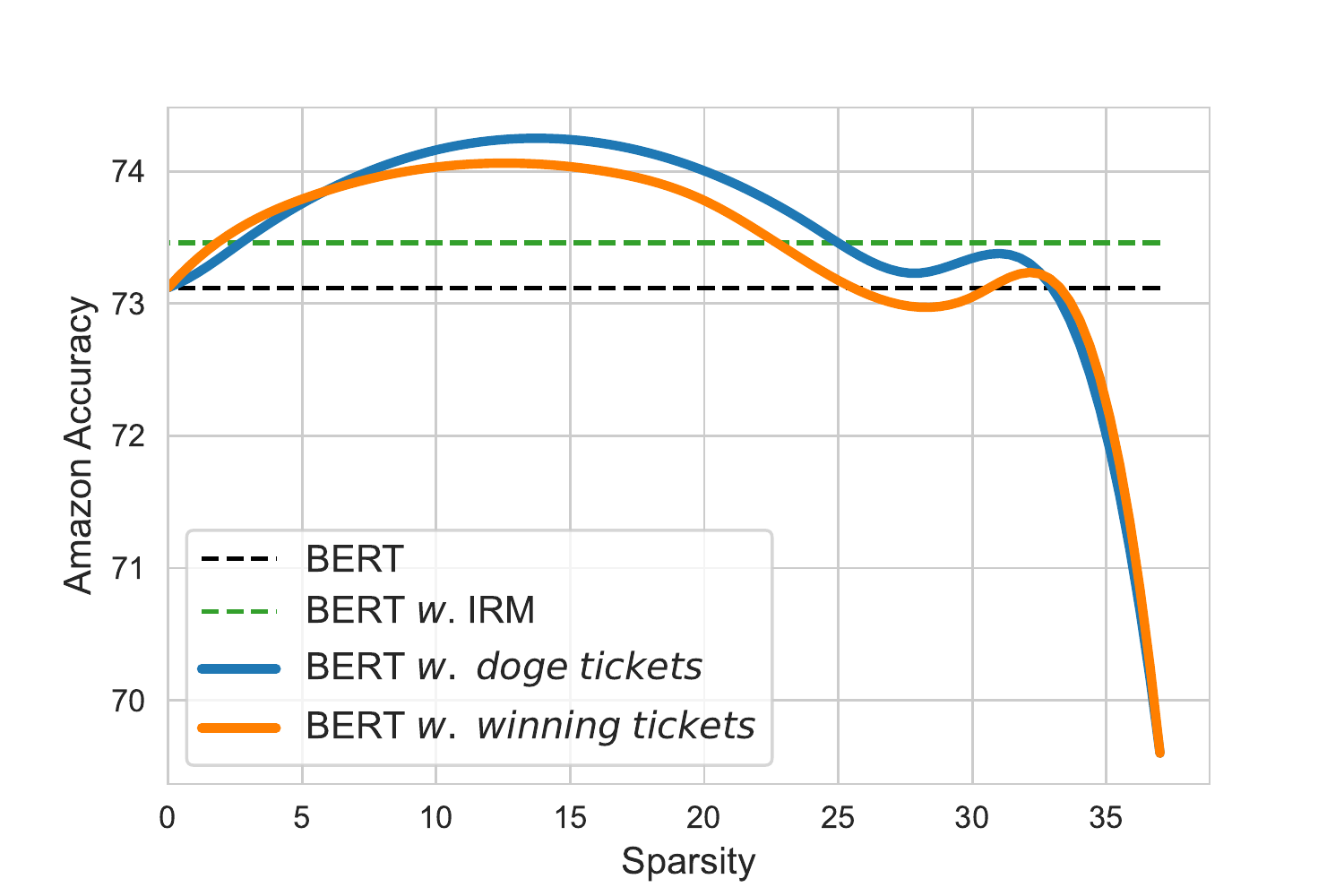}
	\label{fig4b}}\\	
	\caption{Transitions with varying sparsity levels.}
	\label{fig4}
\end{figure}

We prune the parameters under 25 sparsity levels for \texttt{BERT}. While both \textit{doge tickets} and \textit{winning tickets} can be identified at any sparsity levels, the evaluation results of them present transitions when we vary the sparsity levels continuously. We display the transitions of both \texttt{BERT-base} and \texttt{BERT-large} on  \textsc{AmazonA} in Figure~\ref{fig4} to see whether \textit{doge tickets} performs consistently better. 

We can observe that \textit{doge tickets} outperforms \textit{winning tickets} most of the time. Recall the parameter types mentioned in Section~\ref{existence}. At smaller sparsity levels, \textit{doge tickets} tends to first prune LMHV parameters (i.e., domain-specific parameters) instead of LMLV and LMHV parameters which are first considered by \textit{winning tickets}. \textit{doge tickets} shall preserve more domain-general parameters than winning tickets for a better out-of-domain generalization. On the other end, at the larger sparsity levels, \textit{winning tickets} and \textit{doge tickets} start to prune expressive parameters. It can occur that HMLV parameters (i.e., domain-general parameters) will be pruned earlier than HMHV for \textit{winning tickets}. Contrarily, \textit{doge tickets} will certainly prune HMLV parameters later whenever possible. Hence, compared to \textit{winning tickets}, \textit{doge tickets} consistently makes LMs packed with invariant parameters at all sparsity levels to maintain a good out-of-domain generalization.

\section{Conclusions}

In this paper, we address the issue of learning variance of over-parameterized LMs, which we find through a pilot study can lead a critical portion of parameters to behave unexpectedly in a domain-specific manner. Neurons corresponding to these domain-specific parameters are particularly activated for some domains, yet not so active in others. Consequently, LMs are more likely to suffer from a low out-of-domain generalization, as it requires LMs to perform well on different domains. 

Motivated by this observation, we posit that the other parameters are domain-general, i.e., they are applicable across domains, essentially underpinning a domain-general LM that can appropriately derived from the original LM. To uncover the domain-general LM, we have proposed to identify domain-general parameters by playing lottery tickets under the vision of our proposed domain-general score (\textit{doge tickets}). By taking into consideration both the mean and variance of parameter expressiveness, \textit{doge tickets} shows advantages over previous \textit{winning tickets} on the out-of-domain datasets constructed from \textsc{Amazon} and \textsc{Mnli}. Further analyses verify the existence of domain-general parameters and performance consistency of \textit{doge tickets}.

We have empirically shown pruning can help improve out-of-domain generalization of LMs at large. In the future, we plan to examine the maximum potential of pruning by applying it under 1) unsupervised and 2) multi-task scenarios.

\section*{Acknowledgements}

Thank all the anonymous reviewers and chairs for their constructive suggestions.

\bibliography{anthology,custom}
\bibliographystyle{acl_natbib}

\appendix

\begin{figure*}[ht]
    \centering
    \includegraphics[width=0.77\textwidth]{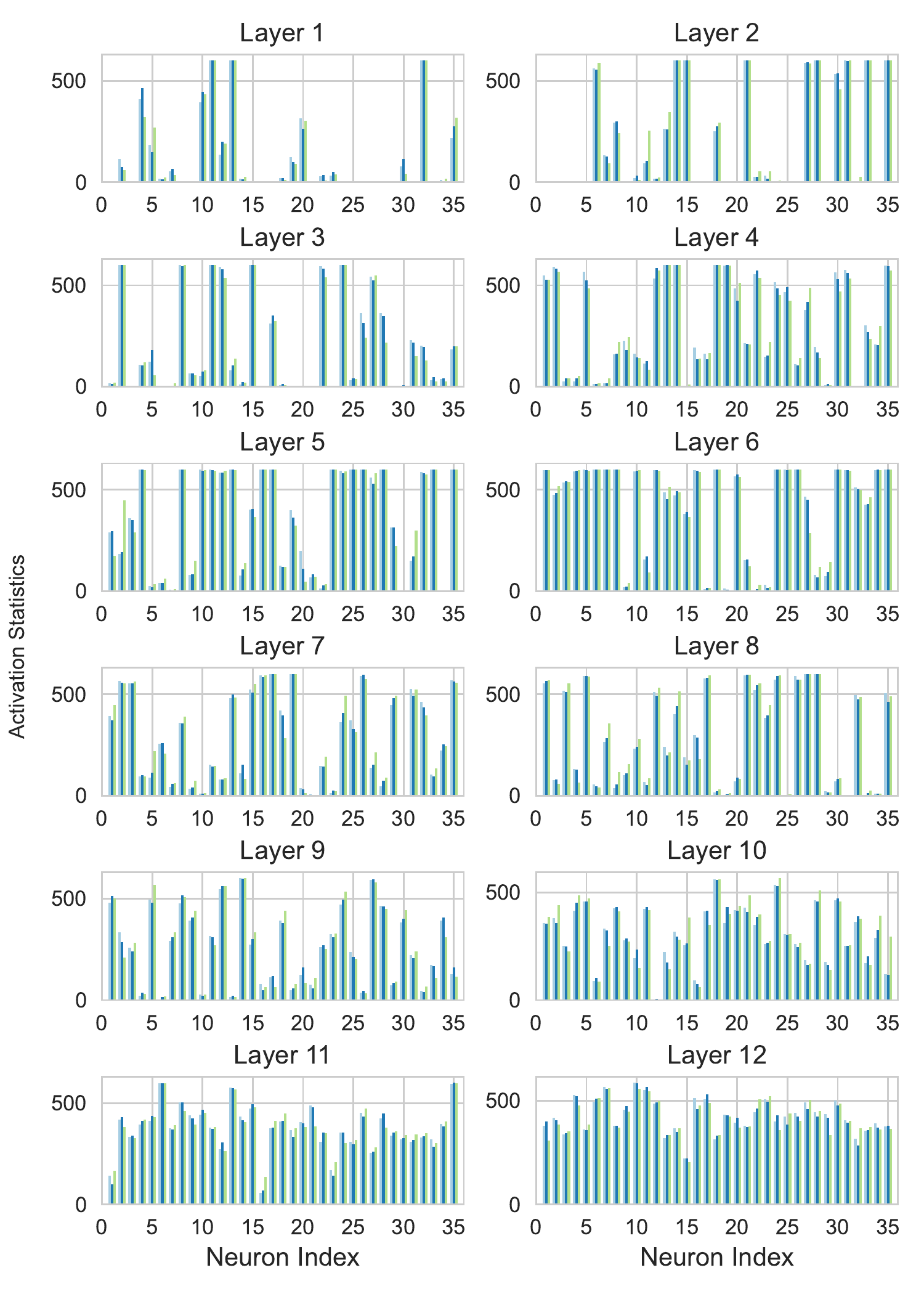}
    \caption{Extended results of our pilot study. Instead of randomly sampled neurons, the first 35 neurons are used.}
    \label{fig6}
\end{figure*}

\section{Extended Results of Our Pilot Study}
\label{appa}

We here present extended results of our pilot study for a better sense of how neurons are activated at different FFN blocks in Figure~\ref{fig6}.

Likewise, domain-specific and domain-general manners can be apparently observed.


\end{document}